\documentclass[10pt,twocolumn,letterpaper]{article}

\usepackage{wacv}
\usepackage{times}
\usepackage{epsfig}
\usepackage{graphicx}
\usepackage{amsmath}
\usepackage{amssymb}
\usepackage{subfigure}
\usepackage{float}
\usepackage{marvosym}

\def\wacvPaperID{708} 

\wacvfinalcopy 

\ifwacvfinal
\fi

\ifwacvfinal
\usepackage[breaklinks=true,bookmarks=false]{hyperref}
\else
\usepackage[pagebackref=true,breaklinks=true,colorlinks,bookmarks=false]{hyperref}
\fi

\begin{document}

\title{Legacy Photo Editing with Learned Noise Prior}

\author{Yuzhi Zhao$^{1}$\textsuperscript{(\Letter)} \quad
Lai-Man Po$^{1}$ \quad  
Tingyu Lin$^{1}$ \quad
Xuehui Wang$^{2}$ \quad
Kangcheng Liu$^{3}$ \\
Yujia Zhang$^{1}$ \quad
Wing-Yin Yu$^{1}$ \quad
Pengfei Xian$^{1}$ \quad
Jingjing Xiong$^{1}$ \\
{\tt\small yzzhao2-c@my.cityu.edu.hk}
\and
$^{1}$City University of Hong Kong \quad
$^{2}$Sun Yat-sen University \quad
$^{3}$The Chinese University of Hong Kong
}


\maketitle

\pagestyle{empty}  
\thispagestyle{empty} 

\begin{abstract}

There are quite a number of photographs captured under undesirable conditions in the last century. Thus, they are often noisy, regionally incomplete, and grayscale formatted. Conventional approaches mainly focus on one point so that those restoration results are not perceptually sharp or clean enough. To solve these problems, we propose a noise prior learner NEGAN to simulate the noise distribution of real legacy photos using unpaired images. It mainly focuses on matching high-frequency parts of noisy images through discrete wavelet transform (DWT) since they include most of noise statistics. We also create a large legacy photo dataset for learning noise prior. Using learned noise prior, we can easily build valid training pairs by degrading clean images. Then, we propose an IEGAN framework performing image editing including joint denoising, inpainting and colorization based on the estimated noise prior. We evaluate the proposed system and compare it with state-of-the-art image enhancement methods. The experimental results demonstrate that it achieves the best perceptual quality. Please see the webpage \href{https://github.com/zhaoyuzhi/Legacy-Photo-Editing-with-Learned-Noise-Prior}{https://github.com/zhaoyuzhi/Legacy-Photo-Editing-with-Learned-Noise-Prior} for the codes and the proposed LP dataset.

\end{abstract}

\section{Introduction}

Restricted by the imaging technology, it remains incomplete parts and noise in legacy grayscale photos. It is highly challenging to restore them due to the great information loss of real world. Also, there is high demand for high-quality and colorful legacy photos. Recently, as deep learning techniques have been demonstrated to successfully applied to many low-level computer vision tasks, the legacy photo enhancement becomes possible. In this paper, we would first discover the representation of blind noise from legacy images as a prior, and then perform image editing based on the estimated noise prior.

Editing legacy photos is highly challenging since there are multiple degradation types in legacy photos. Firstly, there exist noises with unknown distribution and intensity. The noises may be caused by many reasons such as sensor noise, camera distortion, jpeg compression, preservation technology, etc. However, most of current denoisers \cite{zhang2017beyond, tai2017memnet, zhang2018ffdnet, liu2018multi, gu2019self, liu2020densely} are trained with specific noise models such as Gaussian and Poisson distribution. Directly applying those denoisers to legacy photos cannot well enhance the images \cite{abdelhamed2020ntire}. Secondly, it remains flaws or cracks in legacy photos, which are not global noise but regional artifacts. Moreover, the levels of the artifacts are different for distinct pixels, which are hard to estimate. Finally, the grayscale legacy photos lack of color. Thus, the colorization process is significant to attach vivid colors to them. In conclusion, the pipeline of legacy photo editing can be categorized into three parts: denoising, inpainting, and colorization.

\begin{figure*}[t]
\centering
\includegraphics[width=\linewidth, height=6.1cm]{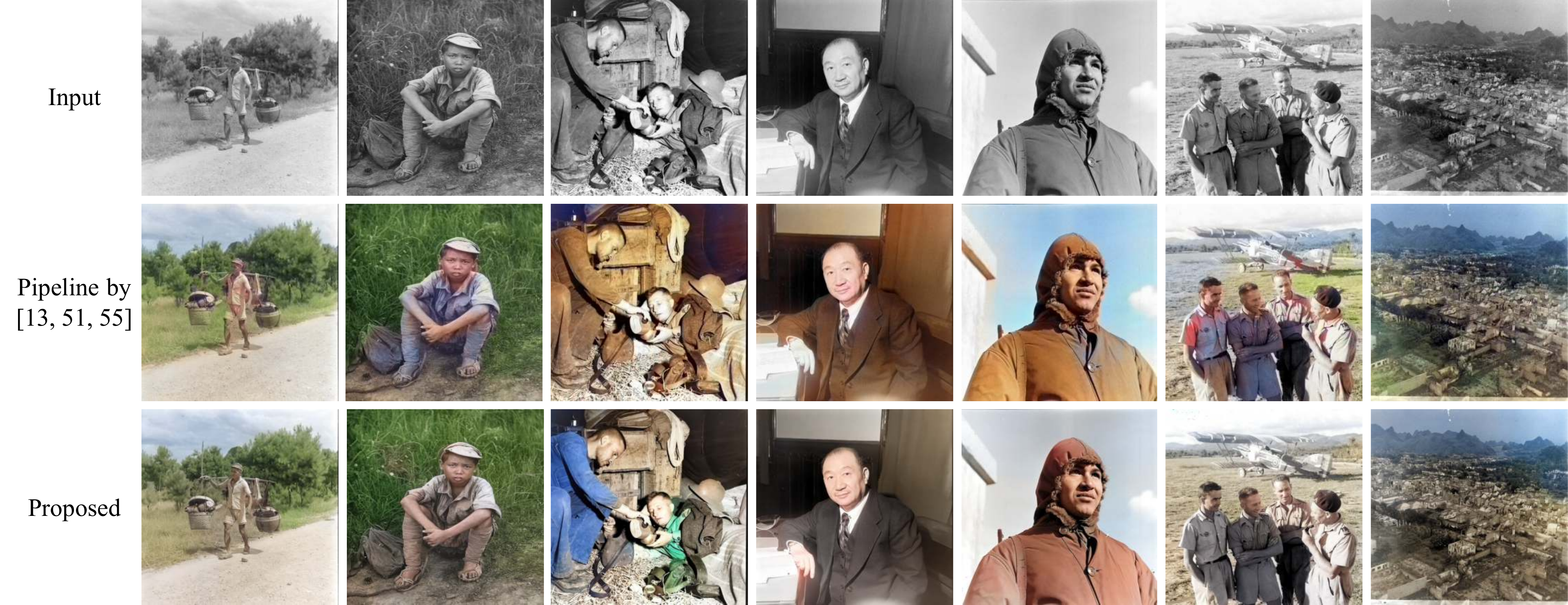}
\caption{The edited real legacy photo samples by proposed method (chosen from LP dataset, captured around 1950). The first, second and third row denote the real legacy photos, image enhancement results by \cite{gu2019self, yu2019free, zhang2016colorful} sequentially and the proposed pipeline, respectively.}
\label{teaser}
\end{figure*}

\begin{figure}[t]
\centering
\includegraphics[width=\linewidth, height=8cm]{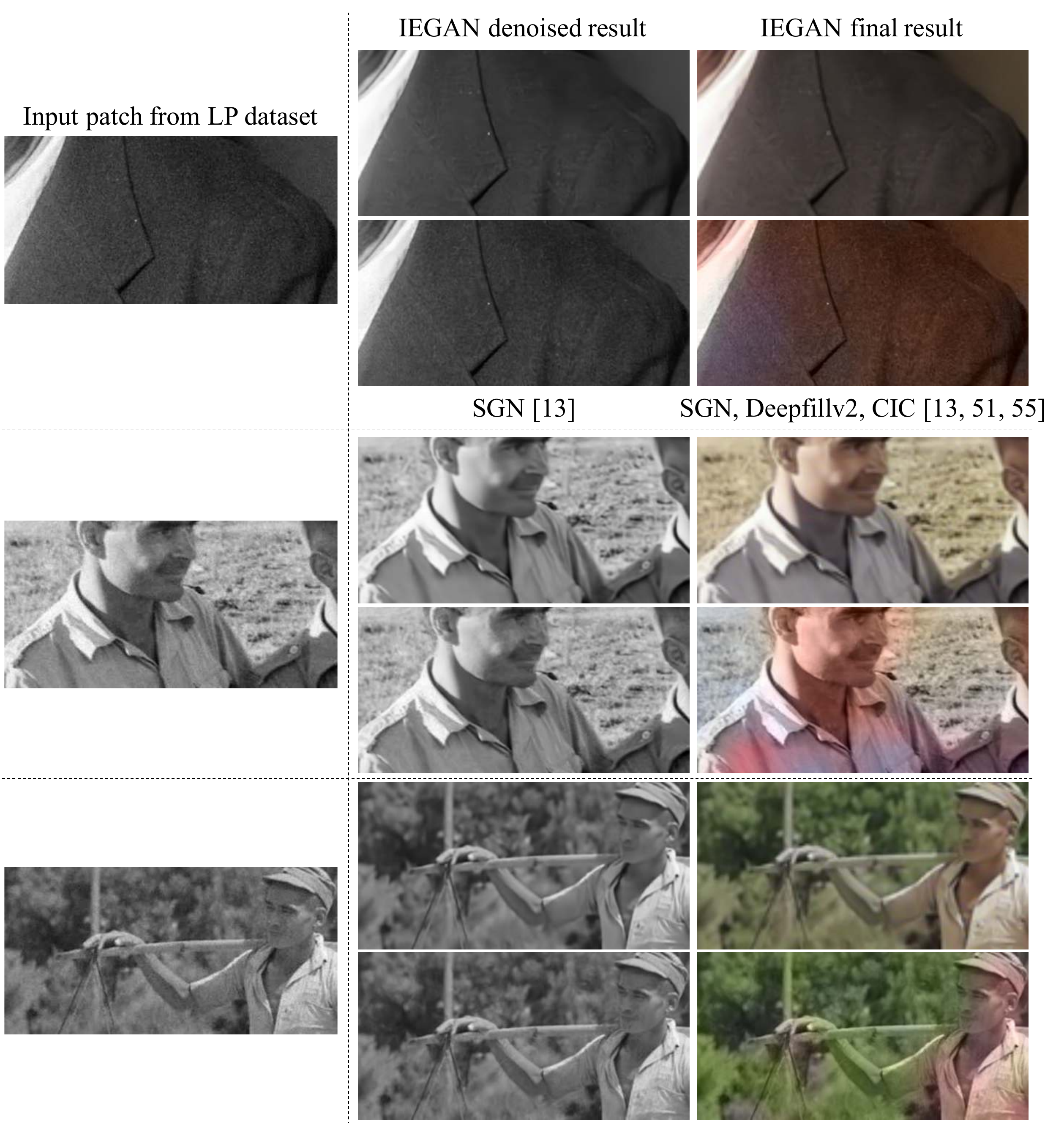}
\caption{The details of edited real legacy photo by 3 samples. The left part includes the input patches. The right part includes denoised result, final result by the proposed system, denoised result by \cite{gu2019self}, and previous pipeline \cite{gu2019self, yu2019free, zhang2016colorful}, respectively.}
\label{teaser}
\end{figure}

To address the issues, we propose a system to implement the pipeline sequentially. Firstly for denoising, the noise distribution of legacy photos is always unknown. However, the current denoisers pre-define a fixed noise model. It is not practical to directly apply the denoisers to process legacy photos with blind noise. If denoised images are not clean enough, the following inpainting and colorization will also be affected. Moreover, there are no pairs of degraded and clean target legacy photos (i.e. legacy photos are normally noisy). There may be three approaches to address the issue such as estimating noise model \cite{abdelhamed2019noise, wei2020physics}, unsupervised training \cite{lehtinen2018noise2noise, batson2019noise2self} and learning blind noise distribution \cite{chen2018image, zamir2020cycleisp}. Since the camera settings are unknown (i.e. the ISP of old cameras is extremely hard to acquire), and the unsupervised training methods also assume a noise distribution, we alternatively propose to learn the noise prior on unpaired legacy photos and clean images by a NEGAN.


Based on the CycleGAN framework \cite{goodfellow2014generative, zhu2017unpaired}, we proposed the NEGAN to estimate the blind noise model. Firstly, we notice that the noisy regions normally include more high-frequency components than common regions; whereas flatten (or noise-free) areas comprise the low-frequency components. Thus, we utilize discrete wavelet transform (DWT) to extract the high-frequency components of generated images and real noisy photos, which are used for computing the domain adversarial loss. Secondly, we randomly select patches rather than resizing whole images, in order to maintain the low-level statistics. In addition, we collect a legacy photo dataset (LP dataset), which contains more than 25000 old photos with different levels of noises.

If the NEGAN is well trained, we can obtain training pairs by manually degrading the clean images from a large-scale dataset, such as ImageNet \cite{russakovsky2015imagenet}. The degraded images have similar statistics with legacy photos. Thus, the following inpainting and colorization processes are based on paired data. The solutions can be briefly categorized into reference-based and automatic. To improve the quality of generated images, we propose to perform the image editing by an IEGAN, i.e. reference-based inpainting and colorization. For the inpainting, it is hard to annotate the real cracks on each legacy photo. Thus, we alternatively collect some templates for modelling the cracks. By multiplying the cracks and clean images, we can obtain the masked images. While for colorization, we use the color scribbles as additional input guidance to enhance colorization reality. The adversarial losses used in IEGAN aim to improve perceptually quality of generated images.

We evaluate the proposed pipeline on ImageNet \cite{russakovsky2015imagenet} validation set. Compared with previous pipelines (i.e. denoising with AWGN, inpainting, and colorization networks), the proposed system achieves the best perceptual performance. Also, we visualize some samples in LP dataset in Figure \ref{teaser} (resolution 1760$\times$1760) and details in Figure \ref{teaser} (resolution 256$\times$512). Since the images in LP dataset generally have little cracks, we manually add the masks to real legacy photos for better visualization.


The main contributions of this paper are as follows:

1) We propose a novel NEGAN for estimating blind noise of legacy photos using unpaired data;

2) We create a new legacy photo dataset (LP dataset) including different types of degradation of real legacy photos for learning noise prior;

3) We propose an IEGAN that jointly performs denoising, inpainting and colorization in a user-guided way based on the noise prior estimated by NEGAN.

\section{Related Work}


\textbf{Image Denoising.} Image denoising is a fundamental problem in low-level vision. Recently, researches have shown that deep learning technologies outperform traditional methods such as bilateral filtering, BM3D \cite{dabov2007image}, non-local algorithm \cite{buades2005non}. Mao et al. \cite{mao2016image} designed U-Net shaped network to perform image denoising, which was improved by DnCNN \cite{zhang2017beyond} using residual learning and MemNet \cite{tai2017memnet} using long memory. Using a tunable noise level map as the input, FFDNet \cite{zhang2018ffdnet} handled a wide range of noise levels and removed spatially variant noise. Considering both Gaussian-Poisson Model and in-camera processing pipeline, CBDNet \cite{guo2019toward} further improved the blind denoising ability by embedding a noise estimation network. To further improve the network architecture, MWCNN \cite{liu2018multi} utilized DWT to avoid down-sampling information loss. SGN \cite{gu2019self} greatly decreased the memory consumption and runtime, while it was further improved by DSWN \cite{liu2020densely} using residual path and reconstruction path.

\textbf{Image Inpainting.} The image inpainting denotes the process of filling cracks of images. Normally, the masks of corresponding masked images are known. Pathak et al. \cite{pathak2016context} firstly adopted a conditional GAN \cite{goodfellow2014generative} for context completion. It was enhanced by jointly utilizing global and local discriminators by Iizuka et al. \cite{iizuka2017globally} to strengthen sharpness for filled regions. Liu et al. \cite{liu2018image} introduced a partial convolution with automatically updated status to deal with irregular input masks. It was improved by gated convolution \cite{yu2019free}. It is the combination of vanilla convolution and gate state, which generalizes the partial convolution by a learnable dynamic feature selection mechanism. The EdgeConnect \cite{nazeri2019edgeConnect} proposed an edge generator and image completion network to minimize blurry effect. Xiong et al. \cite{xiong2019foreground} further enhanced it for foreground-aware image inpainting.


\textbf{Image Colorization.} The existing colorization methods can be briefly categorized into three classes: scribble-based \cite{levin2004colorization, xu2013sparse, chen2012manifold, zhang2017real}, example-based \cite{ironi2005colorization, reinhard2001color, welsh2002transferring, he2018deep, iizuka2019deepremaster}, and fully-automatic \cite{cheng2015deep, zhang2016colorful, iizuka2016let, deshpande2017learning}. The former two kinds of approaches are user-guided that learn a mapping function to propagate user hint to the grayscale image. Since grayscale images only include the edge information, the results are highly relevant to the reasonability of human hints. On the other hand, fully-automatic algorithms directly solve an end-to-end objective from grayscale images to corresponding color embeddings. Normally, these approaches are trained on a very large dataset, which is essential for the system to exploit necessary information from the large-scale database without any human intervention.

\textbf{Generative Adversarial Network for Image Enhancement.} The image enhancement is a general idea to improve the image quality. It is addressed by a list of sub-tasks including demosaicking \cite{zhao2019saliency, chen2018learning}, deblurring \cite{kupyn2018deblurgan, kupyn2019deblurgan}, super-resolution \cite{wang2018esrgan, zhang2019ranksrgan, Xuehui_2020_ACCV}, etc. The performance of image enhancement has been greatly improved through the data-driven deep learning approaches. Generative adversarial network (GAN), developed by Goodfellow et al. \cite{goodfellow2014generative}, defines a minmax game between generator and discriminator. The goal of generator is producing convincing samples which fool discriminator, so as to distinguish generated samples from ground truth. The first well-known general GAN-based image enhancer is Pix2Pix \cite{isola2017image} that translates the images from two different domains. It was improved by Wang et al. \cite{wang2018high} for processing high-resolution images and Zhu et al. \cite{zhu2017unpaired} for multimodal generation.

\section{Methodology}

\subsection{Problem Formulation}

Suppose the clean images are from the domain $Z$ and legacy noisy images are in domain $N$. The target is to process the legacy photo $n \in N$ and obtain colorful clean image $z \in Z$. The images in both domain $Z$ and $N$ are totally different in terms of noise, content, and color.

However, the spatial pixels of clean image $z$ and legacy photo $n$ are not aligned. To constitute valid training pairs, we propose to decolorize $z$ and add pseudo noise to clean image $x \in X$ to obtain image $\hat{x} \in N$, which exists similar low-level characteristics of $n \in N$. We utilize a neural network $G$ to simulate the blind noise for clean image $x$. This transformation process can be formulated as:

\begin{equation}
\hat{x} = G ( x ).
\label{pf1}
\end{equation}

In order to recover colorful clean image $z$ from the artificial degraded image $\hat{x}$, we summarize the process as denoising, inpainting, and colorization, respectively. Similarly, they are implemented by neural networks due to the highly non-linear process. To simulate random mask of legacy photos, we use several binary mask samples $m$ to process $\hat{x}$ and obtain masked degraded image $\ddot{x} = \hat{x} \times m$. In addition to input image, we provide the masks and color scribbles to edit final colorized image and obtain sound perceptual quality. It can be represented as:

\begin{equation}
z = col ( inp ( den(\ddot{x}) ), s ), 
\label{pf2}
\end{equation}
where $col(*)$, $inp(*)$ and $den(*)$ represent the colorization, inpainting, and denoising operations, respectively. The $s$ is the color 
scribble provided by user. In practice, we combine $den(*)$ and $inp(*)$ into one architecture $C$ to accelerate inference, while $col(*)$ is implemented by network $R$.

\begin{figure*}[t]
\centering
\includegraphics[height=6.7cm, width=\linewidth]{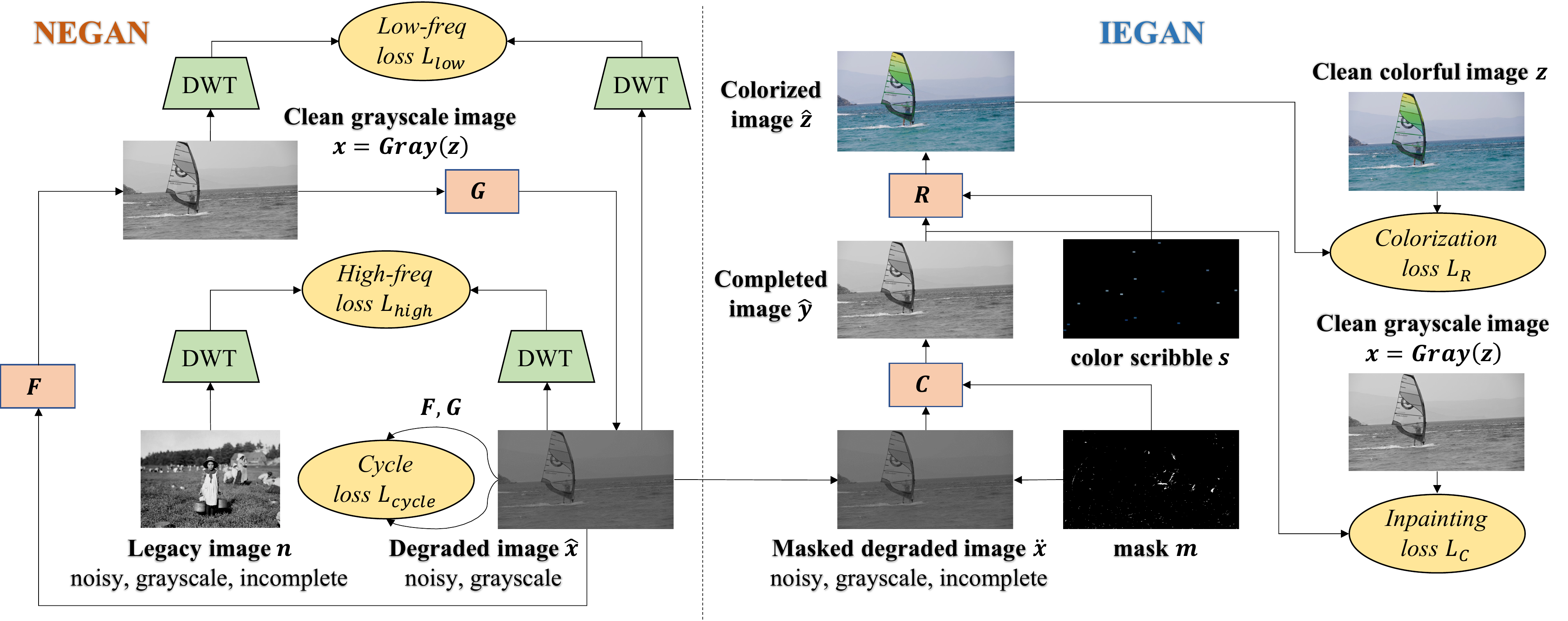} 
\caption{Illustration of training pipeline of proposed method. It contains four convolutional neural networks $G$, $F$, $C$ and $R$. The left part (NEGAN) represents the process that learns noise prior. The right part (IEGAN) shows image editing procedure including the joint denoising, inpainting and scribble-based colorization.}
\label{network}
\end{figure*}


\begin{figure}[t]
\centering
\includegraphics[width=\linewidth]{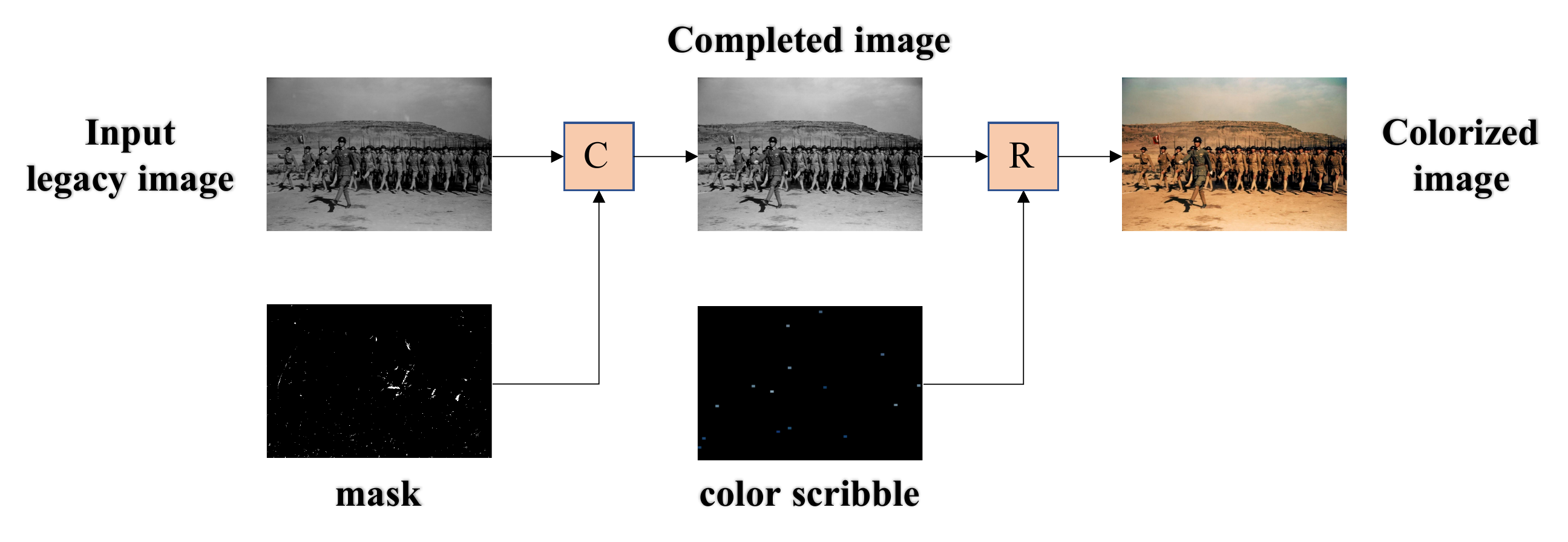}
\caption{Illustration of testing stage of proposed system. The user-provided mask and color scribble map assist the system to produce photorealistic colorizations from legacy photos.}
\label{three}
\end{figure}


\subsection{Training and Testing Pipeline}

Figure \ref{network} shows the training of the proposed pipeline. Specifically, the left and right part of Figure \ref{network} correspond to representations of equation \ref{pf1} and \ref{pf2}, respectively. They are concluded by two architectures, i.e. NEGAN and IEGAN, where NEGAN comprises the sub-network $F$ and $G$, IEGAN comprises the sub-network $C$ and $R$. The NEGAN is trained firstly. Then, it is used to degrade clean images, which are for IEGAN training. To stabilize training of IEGAN sub-networks, we propose to enforce a loss on $C$ directly. Figure \ref{three} shows the testing process, where only $C$, $R$ with additional mask and color scribble are adopted to edit legacy photos. The network architectures and the training details will be presented in following paragraphs.

\subsection{Noise Estimation GAN}

The Noise Estimation GAN (NEGAN) including $G$ and $F$ aims to implement equation \ref{pf1}, i.e. NEGAN translates the images $x \in X$ to noisy image domain $N$. Since there is no evident noise model and paired training data in our application, the unique characteristics of noise become significant. A complete image is composed of low-frequency and high-frequency parts and we notice the noise occupies most high-frequency components of images. Based on the observation, we propose to utilize a clean image $x \in X$ and keep the original low-frequency part $x_L$. Then, we replace its high-frequency component $x_H$ with statistics of noisy image from domain $N$ to implement the translation. Therefore, we propose a Noise Estimation GAN (NEGAN) based on unpaired images to learn the implicit noise distribution, which is called noise prior in following text.

To divide the low-frequency and high-frequency parts, we need to map the images into frequency domain. The common way is to utilize low-pass and high-pass filters, e.g. Gaussian filter and its inverse. It can be defined as:

\begin{equation}
x = x_L + x_H = w \ast x + (\delta - w) \ast x,
\label{pf3}
\end{equation}
where $x$, $w$, $\delta$ represent clean image, low-pass filter, and impulse function, respectively, while ($\delta - w$) is viewed as high-pass filter since it is the reverse of filter $w$. The ``$\ast$'' is convolution operator. But kernel $w$ is often set artificially, which cannot well separate different frequencies. To improve the functionality of the kernel, we introduce discrete wavelet transform (DWT) for frequency division and inverse discrete wavelet transform (IDWT) for image construction. Suppose two components $x_L$ and $x_H$ of input image $x$ are derived from DWT, the whole learning losses for training NEGAN can be represented as:

\begin{equation}
\begin{aligned}
L_{low}(G, F) &= \mathbb{E}[||G(x)_L - x_L||_1] \\ &+ \mathbb{E}[||F(G(x))_L - x_L||_1],
\end{aligned}
\label{negan1}
\end{equation}
\begin{equation}
\begin{aligned}
L_{high}(G, D_N, X, N) &= \mathbb{E}_{x \thicksim X}[||( D_N( G(x)_H ) )^2||] \\ &+ \mathbb{E}_{n \thicksim N}[||( D_N(n_H) - 1 )^2||],
\end{aligned}
\label{negan2}
\end{equation}
\begin{equation}
\begin{aligned}
L_{cycle}(G, F) &= \mathbb{E}_{x \thicksim X}[|| F(G(x)) - x ||_1] \\ &+ \mathbb{E}_{n \thicksim N}[|| G(F(n)) - n ||_1],
\end{aligned}
\label{negan3}
\end{equation}
\begin{equation}
\begin{aligned}
L_{NEGAN} &= \lambda_{low} L_{low}(G, F) + \lambda_{cycle} L_{cycle}(G, F) \\
& + L_{high}(G, D_N, X, N) + L_{high}(F, D_X, X, N),
\end{aligned}
\label{negan4}
\end{equation}
where $G$, $F$, $D_X$, $D_N$ denote generator from domain $X$ to $N$, generator from domain $N$ to $X$, and their corresponding discriminators, respectively. The $x$ and $n$ are random samples from both domains. The $L_{high}$ utilizes the LSGAN loss term \cite{mao2017least}. The $L_{high}$ only matches the low-frequency part of images, which is different from CycleGAN. Also, the discriminators distinguish between fake and real noisy images by matching only the high-frequency part.

\subsection{Image Editing GAN}

The second step of proposed method is to recover a high-quality image from the pseudo noisy image by an Image Editing GAN (IEGAN). The inference of IEGAN is divided into two sub-networks: inpainting network ($C$) and colorization network ($R$). The $C$ generates a complete grayscale image and the $R$ colorizes the output of $C$. As shown in Figure \ref{network}, the proposed IEGAN framework receives pseudo noisy grayscale image with additional mask and color map guidances.

We utilize L1 loss for both sub-networks $C$ and $R$. The losses for them share same representations. It is defined as:

\begin{equation}
L_1 = \mathbb{E}[||t_1 - t_2||_1] ,
\label{iegan1}
\end{equation}
where the two variables $t_1$ and $t_2$ equal to $\hat{y}$ and $x$ for $C$, meanwhile they equal to $\hat{z}$ and $z$ for $R$. The input $\ddot{x} = \hat{x} \odot m$ is a masked grayscale image with an additional Gaussian noise added. The outputs $\hat{y} = C(\ddot{x}, m)$ and $\hat{z} = R(y, s)$. The definitions can be found in Figure \ref{network}.

To boost perceptual quality of generated images, we adopt perceptual loss \cite{johnson2016perceptual}, which is defined as:

\begin{equation}
L_{percep} = \mathbb{E}[||\phi_l (t_1) - \phi_l (t_2)||_1] ,
\label{iegan2}
\end{equation}
where $\phi_l (*)$ represents the features of the $l$-th layer of the pre-trained CNN. In our experiment, we use the $conv_{4\_3}$ layer of VGG-16 \cite{simonyan2014very} network, which is pre-trained on ImageNet \cite{russakovsky2015imagenet} dataset.

Instead of traditional GAN training method \cite{goodfellow2014generative}, we utilize the PatchGAN \cite{isola2017image} with LSGAN critic \cite{mao2017least} to minimize the Pearson $\chi^2$ divergence between the generated samples and ground truth. It is defined as:

\begin{equation}
L_{G} = \cfrac{1}{2} \mathbb{E}[(D(t_1) - 1)^2],
\label{iegan3}
\end{equation}
\begin{equation}
L_{D} = \cfrac{1}{2} \mathbb{E}[(D(t_2) - 1)^2] + \cfrac{1}{2} \mathbb{E}[(D(t_1))^2].
\label{iegan4}
\end{equation}

The total loss functions of IEGAN can be defined as:

\begin{equation}
L_{IEGAN} = L_{1C} + L_{1R} + \lambda_{percep} L_{percepR} + \lambda_{G} L_{GR},
\label{iegan5}
\end{equation}
where inpainting network $C$ only adopts L1 loss term $L_{1C}$. The colorization network $R$ utilizes all three loss terms $L_{1R}$, $L_{percepR}$, and $L_{GR}$. The definitions of the loss terms can also be found in Figure \ref{network}.

\begin{figure}[t]
\centering
\includegraphics[width=\linewidth]{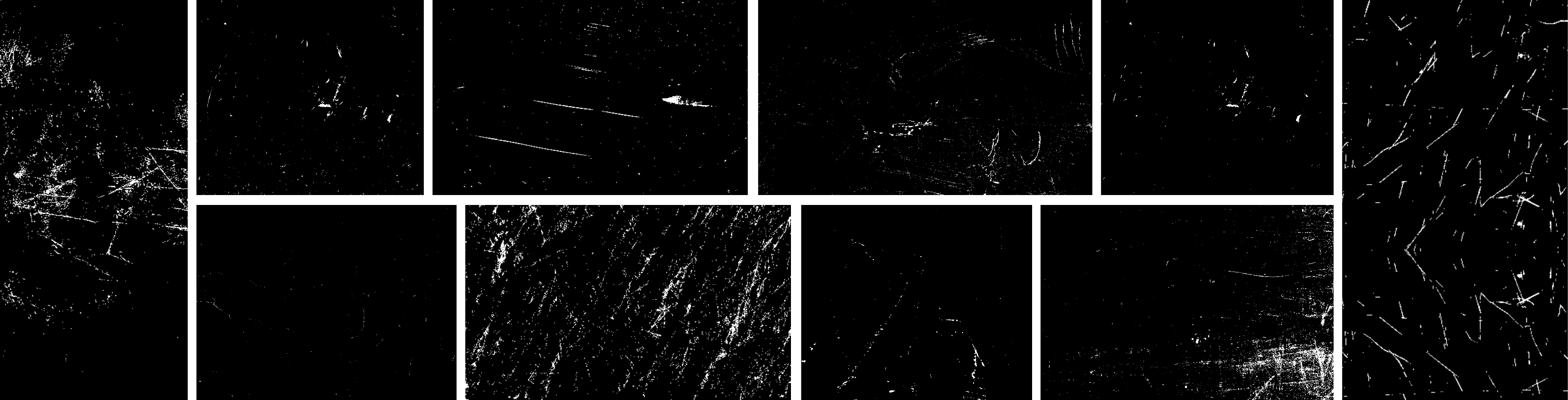}
\caption{Illustration of mask templates used in this paper.}
\label{mask}
\end{figure}

\begin{figure*}[t]
\centering
\includegraphics[width=\linewidth]{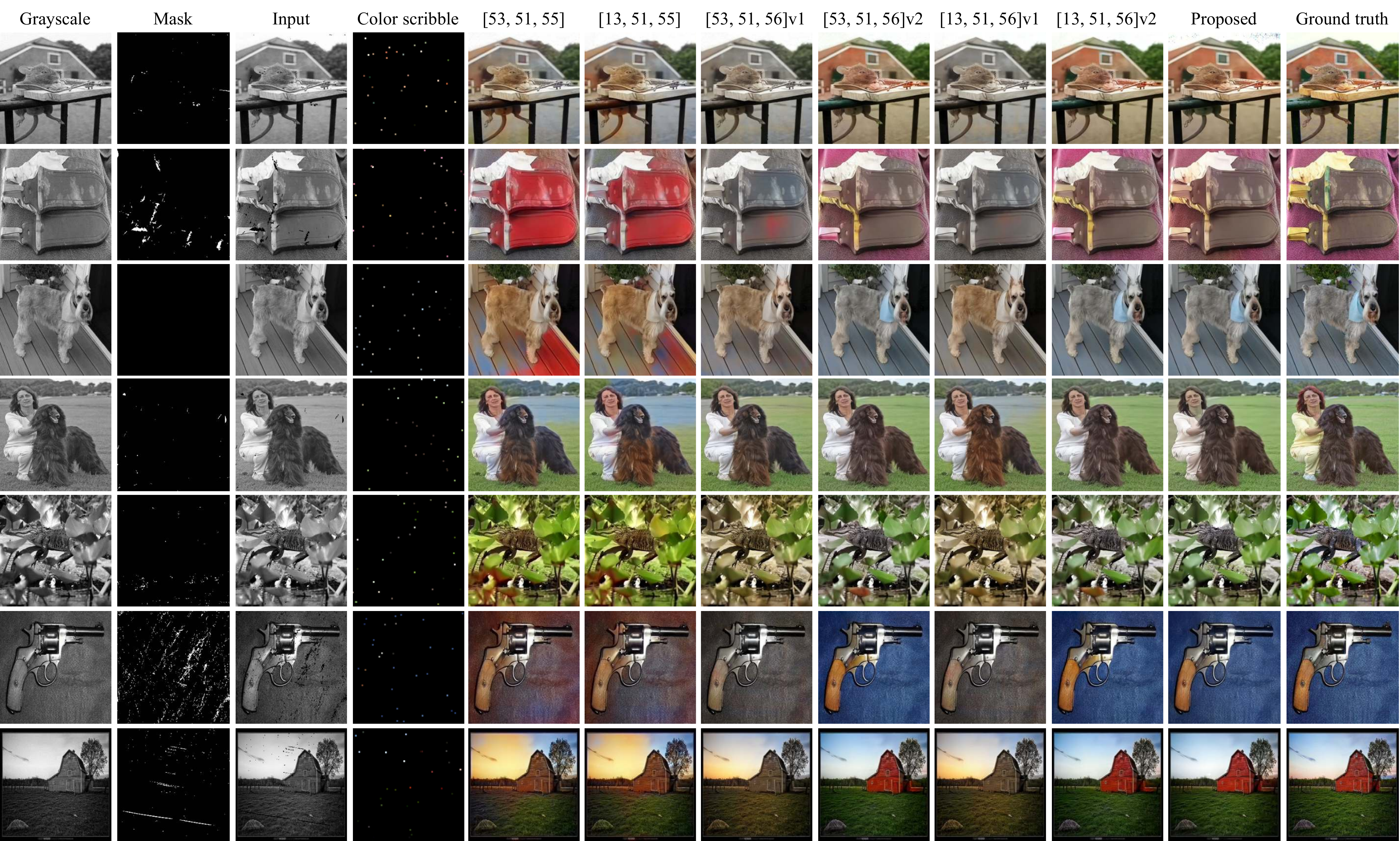}
\caption{Illustration of image editing results. The input masked images are obtained by multiplication operation of grayscale images and masks. Different columns represent different samples edited by methods in experiment. They are randomly selected from validation set.}
\label{compare_fig}
\end{figure*}

\section{Experiment}

\subsection{Implementation Details}

\textbf{Dataset.} We use LP dataset to include enough modes of noisy image domain $N$, for NEGAN training. There are over 25000 grayscale legacy photos with different resolutions in the dataset. Also, we choose ImageNet \cite{russakovsky2015imagenet} (1.3 million images) for clean image domain $X$. It contains 1000 categories, which is general and robust for learning the mapping. At training, we randomly select unpaired sample $n \in N$ and $x \in X$. The images are randomly cropped to 256$\times$256 local patches and normalized to range of [0, 1]. Moreover, the binary mask samples $m$ is randomly cropped from templates, as shown in Figure \ref{mask}.


\textbf{Network Architecture.} For NEGAN architecture, the generators adopt 8 residual blocks \cite{he2016deep} as transformer with residual connection between input and output. There are no downsampling and upsampling operations since they may affect the low-level details. The discriminators adopt $16 \times 16$ PatchGAN architecture and all layers are spectral normalized \cite{miyato2018spectral}. The pre-trained NEGAN produces corresponding degraded images from input while maintains the low-frequency parts. For IEGAN architecture, the generator $C$ and $R$ adopt U-Net structure \cite{ronneberger2015u}. The convolutional layer of $C$ is replaced by gated convolution \cite{yu2019free} to learn adaptive inpainting. The discriminator $D_C$ and $D_R$ adopt convolution part of a VGG-16 architecture while the final output is one channel. The networks are instanced normalized \cite{ulyanov2016instance}. Each layer is LeakyReLU activated \cite{maas2013rectifier}.

\textbf{Optimization.} At first stage, the parameters of all networks are initialized using Xavier method \cite{glorot2010understanding} and the learning rate is initialized as $1 \times 10^{-4}$. The NEGAN and two sub-networks of IEGAN are trained independently for 20 epochs. At second stage, all networks are optimized jointly. The learning rate is fixed to $5 \times 10^{-5}$ while the system is trained for another 20 epochs. The learning rate is fixed in both stages. We use Adam optimizer \cite{kingma2014adam} with $\beta_1$ = 0.5, $\beta_2$ = 0.999 and batch size of 32. Moreover, we randomly select 0 - 30 color scribbles as hint for network $R$. The hyperparameters $\lambda_{percep}, \lambda_{G}$ equal to 10 and 0.1, respectively. We implement our system with PyTorch framework and train it on 4 NVIDIA Titan Xp GPUs. It takes approximately 2 weeks to complete the whole training process.

\begin{table}[t]
\caption{Comparison results of the proposed pipeline and other 6 state-of-the-art pipelines. The grayscale images (clean) are obtained from ground truth colorful images. In ``Reference'' item, the ``mask'' and ``color'' denote the additional mask and color scribble input. Also, the \cite{zhang2017beyond,yu2019free,zhang2016colorful} represents using \cite{zhang2017beyond}, \cite{yu2019free}, \cite{zhang2016colorful} for inference sequentially.}
\begin{center}
\begin{tabular}{lccc}
\hline
Method & Reference & PSNR & SSIM \\
\hline
\hline
Grayscale (clean) & / & 23.24 & 0.9443 \\
\hline
\cite{zhang2017beyond,yu2019free,zhang2016colorful} & mask & 21.26 & 0.8865 \\
\hline
\cite{gu2019self,yu2019free,zhang2016colorful} & mask & 21.18 & 0.8865 \\
\hline
\cite{zhang2017beyond,yu2019free,zhang2017real}v1 & mask & 23.62 & 0.9059 \\
\hline
\cite{zhang2017beyond,yu2019free,zhang2017real}v2 & mask, color & 27.51 & 0.9233 \\
\hline
\cite{gu2019self,yu2019free,zhang2017real}v1 & mask & 23.50 & 0.9024 \\
\hline
\cite{gu2019self,yu2019free,zhang2017real}v2 & mask, color & 27.34 & 0.9194 \\
\hline
Proposed & mask, color & \textbf{28.02} & \textbf{0.9408} \\
\hline
\end{tabular}
\end{center}
\label{comparison}
\end{table}


\subsection{Validation on Image Editing Quality}

In this section, we quantitatively evaluate the image enhancement quality of the proposed system. Since there is no ground truth for legacy photos, we alternatively adopt the ImageNet validation 50000 images. We convert the images to grayscale and rescale them to 256$\times$256. Each validation image is added a pseudo mask and an additive Gaussian noise with standard deviation of 0.05 to simulate a legacy image, which is similar to training process. At inference stage, only IEGAN is used since the noise prior modelled by NEGAN is implied in $C$ at training. We utilize different combinations of denoisers \cite{zhang2017beyond, gu2019self}, inpainting network \cite{yu2019free}, and colorization networks \cite{zhang2016colorful,zhang2017real} as pipelines and there are overall 6 combinations. All aforementioned algorithms are trained on ImageNet training dataset. Specifically, the denoisers are trained on the same noise level (i.e. AWGN) as validation data, whereas IEGAN is trained on blind noise learned from noise prior. The method \cite{zhang2017real} is a scribble-based colorization algorithm while \cite{zhang2016colorful} is fully-automatic colorization method. Color scribbles are used in both IEGAN and method \cite{zhang2017real}; therefore all the approaches in experiment adopt reference information. There are 30 scribbles used for IEGAN and for \cite{zhang2017real} at test.

The comparison results are summarized in Table \ref{comparison} and illustrated in Figure \ref{compare_fig}. The methods using color scribbles achieve the PSNR higher than 27 and obviously outperform others since they have precise color prior. Note that, the proposed method achieves the highest PSNR and SSIM since the two parts of IEGAN are trained collaboratively. Therefore, the colorized images are more natural and realistic than other methods.

\begin{table}[t]
\caption{Comparison results for ablation study.}
\begin{center}
\begin{tabular}{lccc}
\hline
Methods & Setting & PSNR & SSIM \\
\hline
\hline
w/o DWT-based losses & 1) & 27.91 & 0.9397 \\
\hline
w/o perceptual loss & 2) & 27.78 & 0.9369 \\
\hline
w/o GAN loss & 2) & 27.83 & 0.9396 \\
\hline
w/o both losses & 2) & 26.18 & 0.9334 \\
\hline
10 color scribbles & 3) & 26.78 & 0.9354 \\
\hline
20 color scribbles & 3) & 27.59 & 0.9390 \\
\hline
w/o joint training & 4) & 27.45 & 0.9358 \\
\hline
Proposed & / & \textbf{28.02} & \textbf{0.9408} \\
\hline
\end{tabular}
\end{center}
\label{ablation}
\end{table}



\begin{figure*}[t]
\centering
\includegraphics[width=\linewidth]{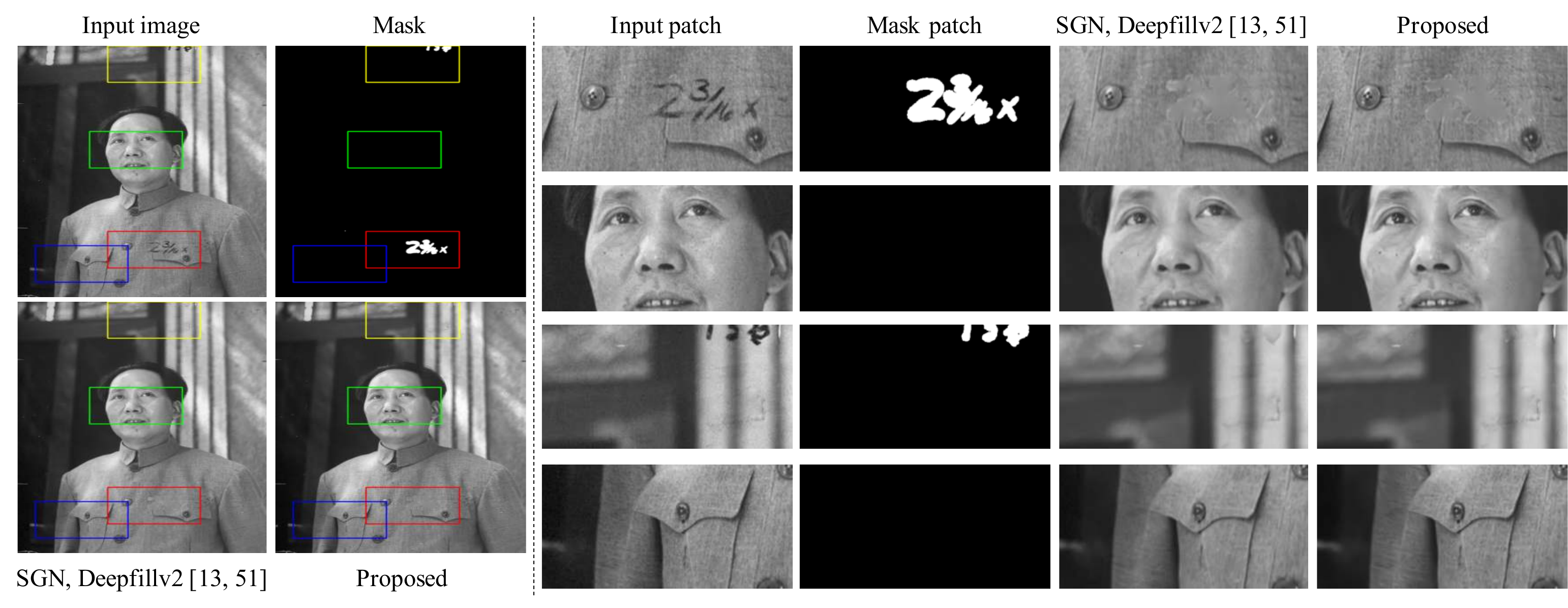}
\caption{Comparison of legacy photo enhancement results of the proposed and previous \cite{gu2019self, yu2019free} pipeline. The left part and right part include the full resolution legacy photos and local patches, respectively. The colorful rectangles denote the locations of selected patches.}
\label{denoised}
\end{figure*}

\subsection{Ablation Study}

In order to demonstrate the effectiveness of NEGAN and IEGAN losses, we set up 4 ablation study settings. We use 50000 ImageNet validation data for validation. All images are added unknown noise by pre-trained NEGAN to simulate legacy photos. The settings are show as:

1) Drop the DWT-based loss terms that NEGAN noise prior learner retrogrades to a CycleGAN \cite{zhu2017unpaired};

2) Drop the perceptual loss or GAN loss or both loss terms of IEGAN to compare their effectiveness, while the NEGAN remains unchanged;

3) Decrease the number of color scribbles to 20 or 10;

4) Train two sub-networks of IEGAN framework separately in order to evaluate joint training scheme.

As shown in Table \ref{ablation}, the full system reaches the best performance on PSNR and SSIM \cite{wang2004image}. If the DWT-based loss terms are dropped, the system is hard to handle the ``real noise'' generated by the NEGAN. Also, each loss term or joint training contributes to better performance. In conclusion, all components of proposed method and significant.


\begin{figure*}[t]
\centering
\includegraphics[width=\linewidth]{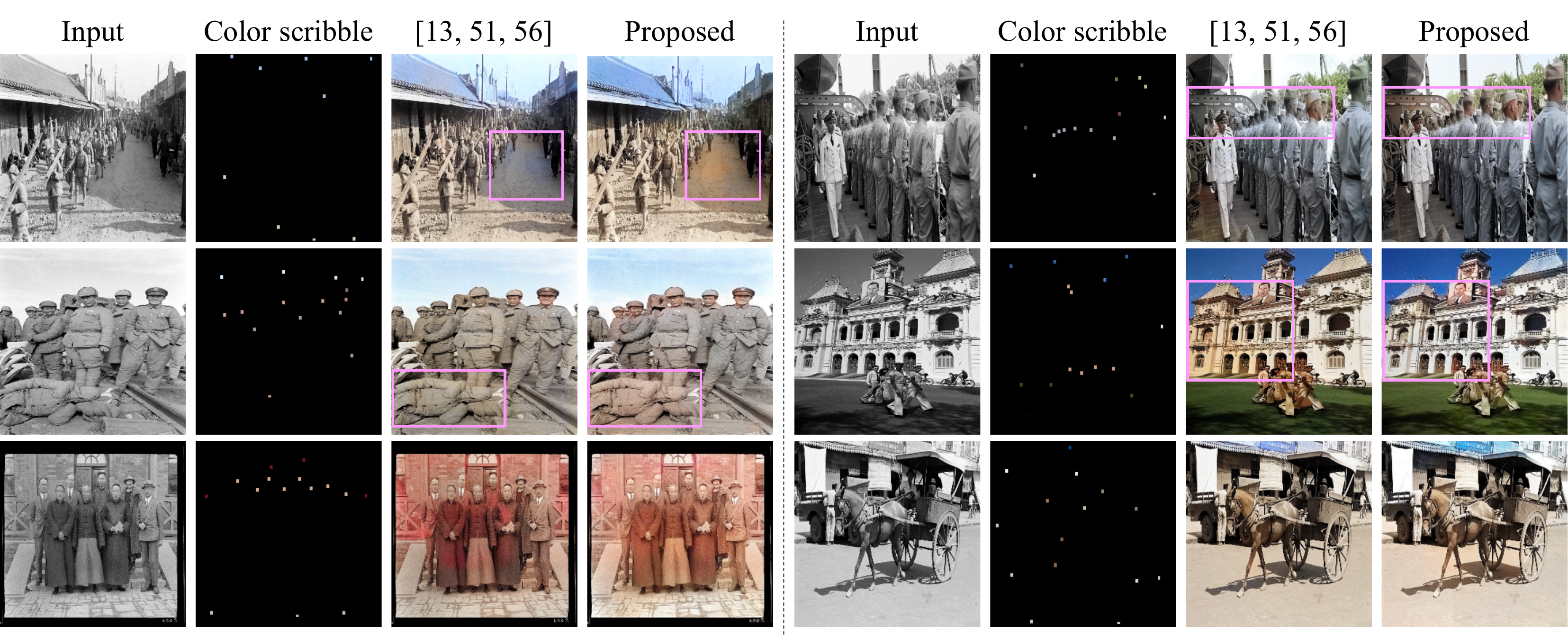}
\caption{Comparison of the proposed and previous \cite{gu2019self, yu2019free, zhang2017real} pipeline on real legacy photos. The rectangles denote the highlighted areas.}
\label{legacy}
\end{figure*}

\subsection{Validation on Legacy Photo Enhancement}

In this section, we assess the denoising and inpainting ability of the proposed system, i.e. network $C$ of IEGAN. The state-of-the-art denoising and inpainting methods \cite{gu2019self, yu2019free} are used for comparison. For the denoising, the results of proposed approach are more sharper than \cite{gu2019self, yu2019free}. For instance, the eyebrows, cheeks and beard generated by the proposed method are more clear, as shown in the second patch in Figure \ref{denoised}. For inpainting, the patches produced by the proposed method are also realistic. For instance, the color of filled regions are closer to clothes, as shown in the first patch. Also, the patch of proposed model in third row is much more smoother than \cite{gu2019self, yu2019free}. Since the NEGAN better estimates the noise model, the generated results are cleaner and sharper. Moreover, the inpainted regions are more plausible due to better denoising ability.

\subsection{Validation on Legacy Photo Colorization}

In this section, we assess the editing quality of the proposed system on real legacy photos. We utilize the state-of-the-art pipelines, i.e. \cite{gu2019self, yu2019free, zhang2017real} for comparison and 15 color scribbles are adopted for both methods, as shown in Figure \ref{legacy}. The samples are selected from the proposed LP dataset. The proposed method produces more plausible colors than compared method since the it utilizes joint training scheme for image denoising, inpaiting and colorization. Moreover, the proposed method learns noise prior well, thus it produces high-quality images.

\begin{figure*}[t]
\centering
\includegraphics[width=\linewidth]{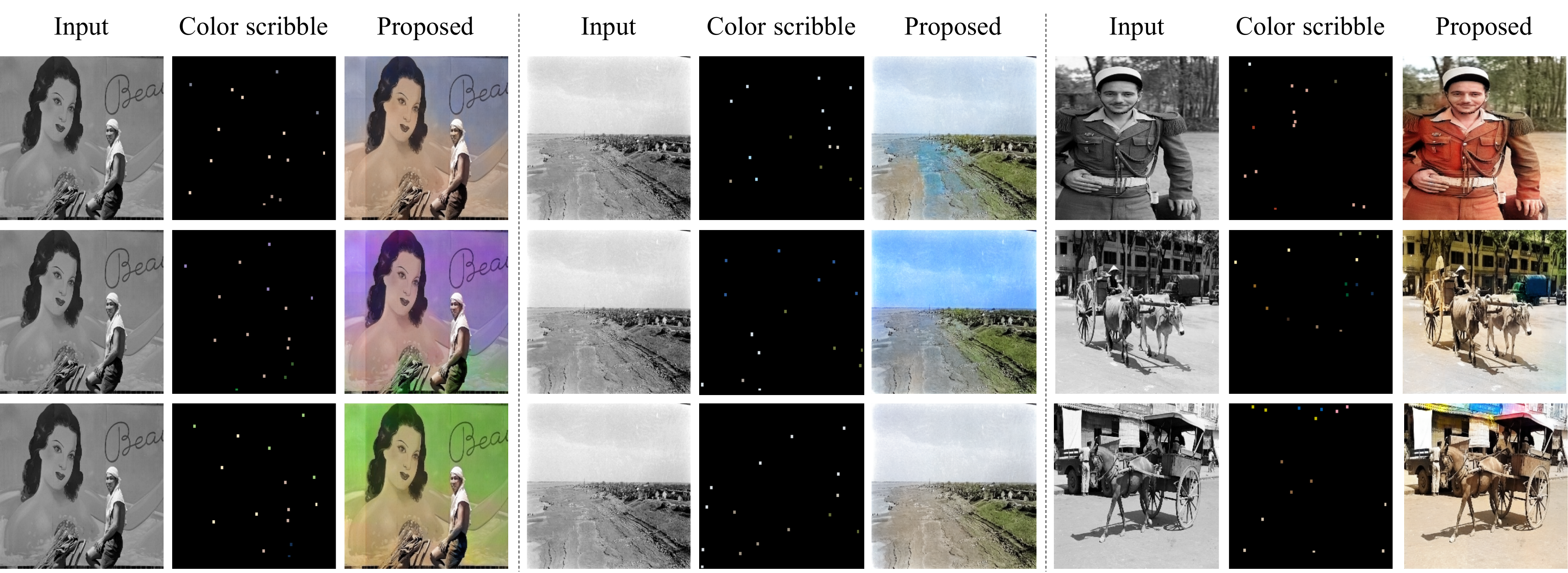}
\caption{Illustration of some failure image editing cases of proposed method, including color bleeding, artifacts and inconsistent colors.}
\label{failure}
\end{figure*}

\section{Failure Cases}

Our system can predict relatively reasonable colorizations in many cases; however, there are still some common failure cases, shown in Figure \ref{failure}. For left part (left 3 rows in the figure), there exists slight color bleeding effect when given “not reasonable color scribbles”. Since the degradation degree is shifted for many legacy photos, the output images of center part still contain artifacts. Finally, as color scribbles provided by users are not reasonable enough, the results are also not very plausible, as right part shows. We will enhance the design of the proposed framework and. Moreover, we will add semantic information into our framework to guide inpainting and colorization in the future.

\section{Conclusion}

In this paper, we present a novel framework for editing legacy photos in an end-to-end manner. Since the legacy photographs are captured by old cameras, they are corrupted with undesirable noise, artifacts and saved in grayscale format. The noise is often blind, thus it is difficult to use a specific distribution for modelling. Thus, we propose a NEGAN to simulate noise prior learned from real legacy photos based on unpaired data training. We enforce the NEGAN to focus more on noisy parts (i.e. high-frequency components) of images by introducing DWT-based loss functions. Moreover, we collect a large-scale legacy photo dataset (LP dataset) including more than 25000 real photographs in different scenes for training NEGAN. Moreover, to remove the artifacts and colorize legacy photos, we propose an IEGAN that performs joint denoising, inpainting and scribble-based colorization sequentially, based on estimated noise prior. At test phase, users can edit the legacy photo by providing masks and color scribbles. Experimental results show that the proposed framework has better performance than the state-of-the-art pipelines.

\section*{Acknowledgement}

The authors would like to thank the colleagues of City University of Hong Kong for their support for collecting legacy photos. The authors would also like to thank the anonymous reviewers and conference chairs for their kind suggestions.

{\small
\bibliographystyle{ieee_fullname}
\bibliography{0708}
}

\end{document}